    \documentclass[10pt,twocolumn,letterpaper]{article}
    \usepackage[pagenumbers]{cvpr} 
    \usepackage{graphicx}
    \usepackage{amsmath}
    \usepackage{amssymb}
    \usepackage{booktabs}
    
    \usepackage[pagebackref,breaklinks,colorlinks]{hyperref}
    \usepackage{multirow}
    
    \usepackage[dvipsnames]{xcolor}
    \newcommand{\tabincell}[2]{\begin{tabular}{@{}#1@{}}#2\end{tabular}}
    \newcommand\blfootnote[1]{%
      \begingroup
      \renewcommand\thefootnote{}\footnote{#1}%
      \addtocounter{footnote}{-1}%
      \endgroup
    }
    \definecolor{violet}{rgb}{0.56, 0.0, 1.0}
    
    \usepackage[pagebackref,breaklinks,colorlinks]{hyperref}
    
    \usepackage[capitalize]{cleveref}
    \crefname{section}{Sec.}{Secs.}
    \Crefname{section}{Section}{Sections}
    \Crefname{table}{Table}{Tables}
    \crefname{table}{Tab.}{Tabs.}
    
    \def\cvprPaperID{5612} 
    \def\confName{CVPR}
    \def\confYear{2022}

    \usepackage{subfiles}
    
\begin{document}
    
    \title{
    Dual-AI: Dual-path Actor Interaction Learning for Group Activity Recognition
    }
    
    \author{Mingfei Han$^{*1}$, David Junhao Zhang$^{*2}$, Yali Wang$^{*3}$, Rui Yan$^{2}$, 
    Lina Yao$^{5}$,\\ Xiaojun Chang$^{1,4}$, Yu Qiao$^{\dagger 3,6}$\\ 
      $^1$ReLER, AAII, UTS \quad
      $^2$National University of Singapore \\
      $^3$ShenZhen Key Lab of Computer Vision and Pattern Recognition,
      SIAT-SenseTime Joint Lab, \\
      Shenzhen Institutes of Advanced Technology, Chinese Academy of Sciences \quad
      $^4$RMIT University \\
      $^5$University of New South Wales\quad
      $^6$Shanghai AI Laboratory, Shanghai, China\\
    \url{https://mingfei.info/Dual-AI/}
    }
    \maketitle
    
    \blfootnote{$^*$ Equal contribution.\ \  $\dagger$ Corresponding author.}

    \begin{abstract}
        Learning spatial-temporal relation among multiple actors is crucial for group activity recognition. 
        Different group activities often show the diversified interactions between actors in the video. 
        Hence, 
        it is often difficult to model complex group activities from a single view of spatial-temporal actor evolution. 
        To tackle this problem, 
        we propose a distinct Dual-path Actor Interaction (Dual-AI) framework, 
        which flexibly arranges spatial and temporal transformers in two complementary orders, 
        enhancing actor relations by integrating merits from different spatio-temporal paths. 
        Moreover, 
        we introduce a novel Multi-scale Actor Contrastive Loss (MAC-Loss) between two interactive paths of Dual-AI. 
        Via self-supervised actor consistency in both frame and video levels, 
        MAC-Loss can effectively distinguish individual actor representations to reduce action confusion among different actors.
        Consequently, 
        our Dual-AI can boost group activity recognition by fusing such discriminative features of different actors. 
        To evaluate the proposed approach, 
        we conduct extensive experiments on the widely used benchmarks, 
        including Volleyball \cite{ibrahim2016hierarchical}, 
        Collective Activity \cite{choi2009they}, 
        and NBA datasets \cite{yan2020social}. 
        The proposed Dual-AI achieves state-of-the-art performance on all these datasets. 
        It is worth noting the proposed Dual-AI with 50\% training data outperforms a number of recent approaches with 100\% training data.
        This confirms the generalization power of Dual-AI for group activity recognition, 
        even under the challenging scenarios of limited supervision.
        \end{abstract}
    
    \begin{figure}[t]
       \begin{center}
       \includegraphics[width=1.0\linewidth]{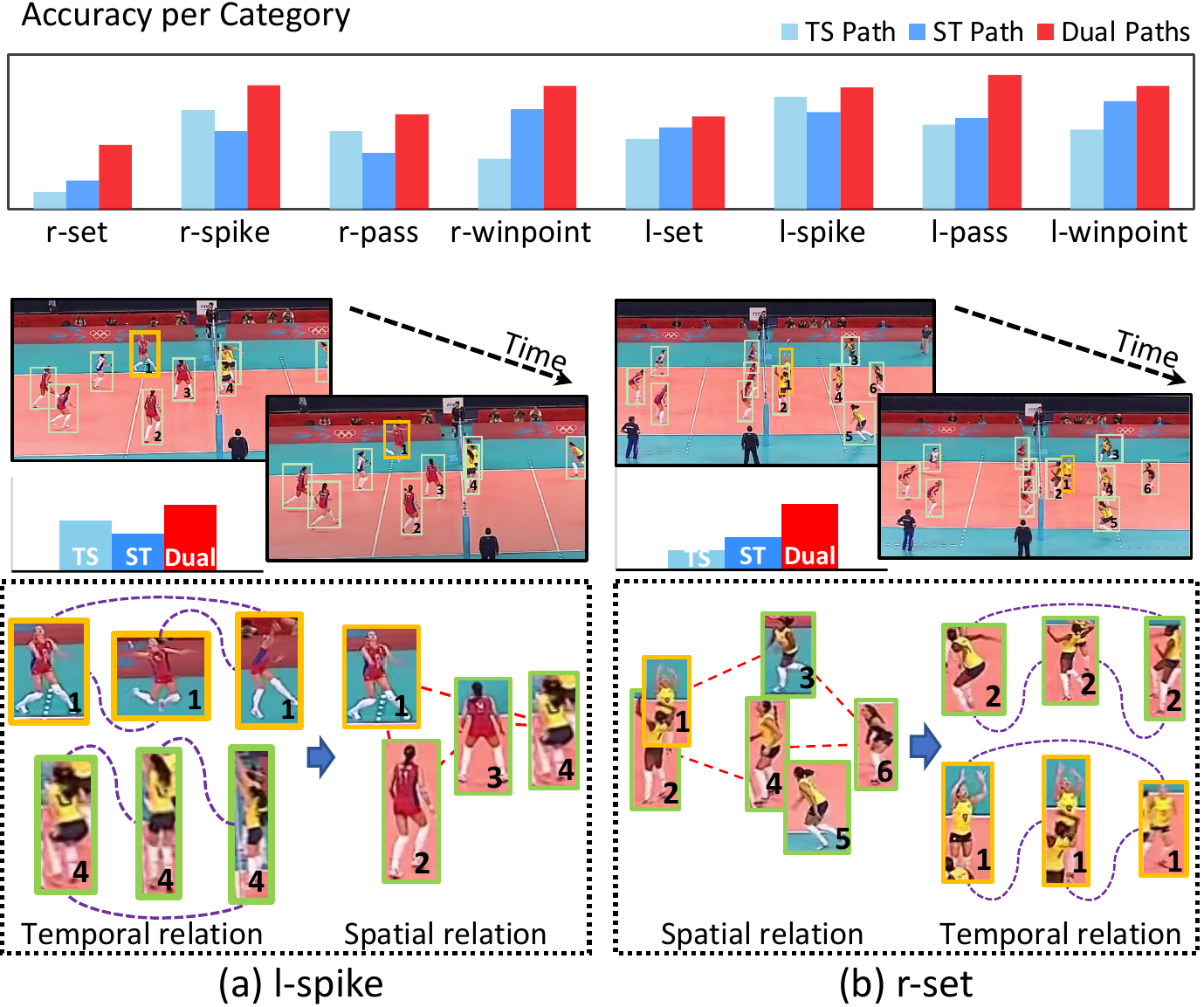}
       \end{center}
          \vspace{-4mm}
          \caption{
          Accuracy per Category and Example of \textit{left spike} and \textit{right set} group activity.
          \textcolor{red}{Red} dashed line and \textcolor{violet}{Violet} dashed line below show spatial and temporal actor interaction respectively.
          With spatial and temporal modeling applied in different orders,
          ST path and TS path learn different spatiotemporal patterns and thereby are skilled at different classes,
          supported by the accuracy plot.
         }
         \label{fig:compl_pattern}
         \vspace{-4mm}
    \end{figure}
    
    \section{Introduction}
       \label{sec:intro}
      
        Group Activity Recognition (GAR) is an important problem in video understanding.
        In this task,
        we should not only recognize individual action of each actor but also understand collective activity of multiple involved actors.
        Hence,
        it is vital to learn spatio-temporal actor relations for GAR.
       
        Several attempts have been proposed to model actor relations by building visual attention among actors \cite{hu2020progressive,wu2019learning,yan2020social,yuan2021spatio,gavrilyuk2020actor,li2021groupformer,azar2019convolutional}.
        However,
        it is often difficult for joint spatial-temporal optimization \cite{tran2018closer,bertasius2021space}. 
        For this reason,
        the recent approaches in group activity recognition often decompose spatial-temporal attention separately for modeling actor interaction \cite{gavrilyuk2020actor,li2021groupformer,yan2020social}.
        But single order of space and time is insufficient to describe complex group activities,
        due to the fact that different group activities often exhibit diversified spatio-temporal interactions.
    
        For example,
        \cref{fig:compl_pattern} (a) refers to the \textit{l-spike} activity in the volleyball,
        where
        the hitting player (actor 1) and the defending player (actor 4) move fast to hit and block the ball,
        while other accompanying players (\eg, actor 2 and actor 3) stand without much movement.
        Hence,
        for this group activity,
        it is better to first understand temporal dynamics of each actor,
        and then reason spatial interaction among actors in the scene.
        On the contrary,
        \cref{fig:compl_pattern} (b) refers to the \textit{r-set} activity in the volleyball,
        where most players in the right-side team are moving cooperatively to tackle the ball falling on different positions,
        \eg,
        actor 1 jumps and sets the ball,
        while actor 2 jumps together to make a fake spiking action.
        Hence,
        for this group activity,
        it is better to reason spatial actor interaction first to understand the action scene,
        and then model temporal evolutions of each actor.
        In fact,
        as shown in the accuracy plot of \cref{fig:compl_pattern},
        the order of space and time interaction varies for different activity categories.

        Based on these observations,
        we propose a distinct Dual-path Actor Interaction (Dual-AI) framework for GAR,
        which can effectively integrate two complementary spatiotemporal views to learn complex actor relations in videos.
        Specifically,
        Dual-AI consists of Spatial-Temporal (ST) and Temporal-Spatial (TS) Interaction Paths,
        with assistance of spatial and temporal transformers.
        ST path first takes spatial transformer to capture spatial relation among actors in each frame,
        and then utilizes temporal transformer to model temporal evolution of each actor over frames.
        Alternatively,
        TS path arranges spatial and temporal transformers in a reverse order to describe complementary pattern of actor interaction.
        In this case,
        our Dual-AI can comprehensively leverage both paths to generate robust spatiotemporal contexts for boosting GAR.
        
        Furthermore, 
        we introduce a novel Multi-scale Actor Contrastive Loss (MAC-Loss),
        which is a concise but effective self-supervised signal to enhance actor consistency between two paths.
        Via such actor supervision in all the frame-frame, frame-video, video-video levels,
        we can further reduce action confusion between any two individual actors to improve the discriminative power of actor representations in GAR.
        
\begin{figure}[t]
       \begin{center}
       \includegraphics[width=1.\linewidth]{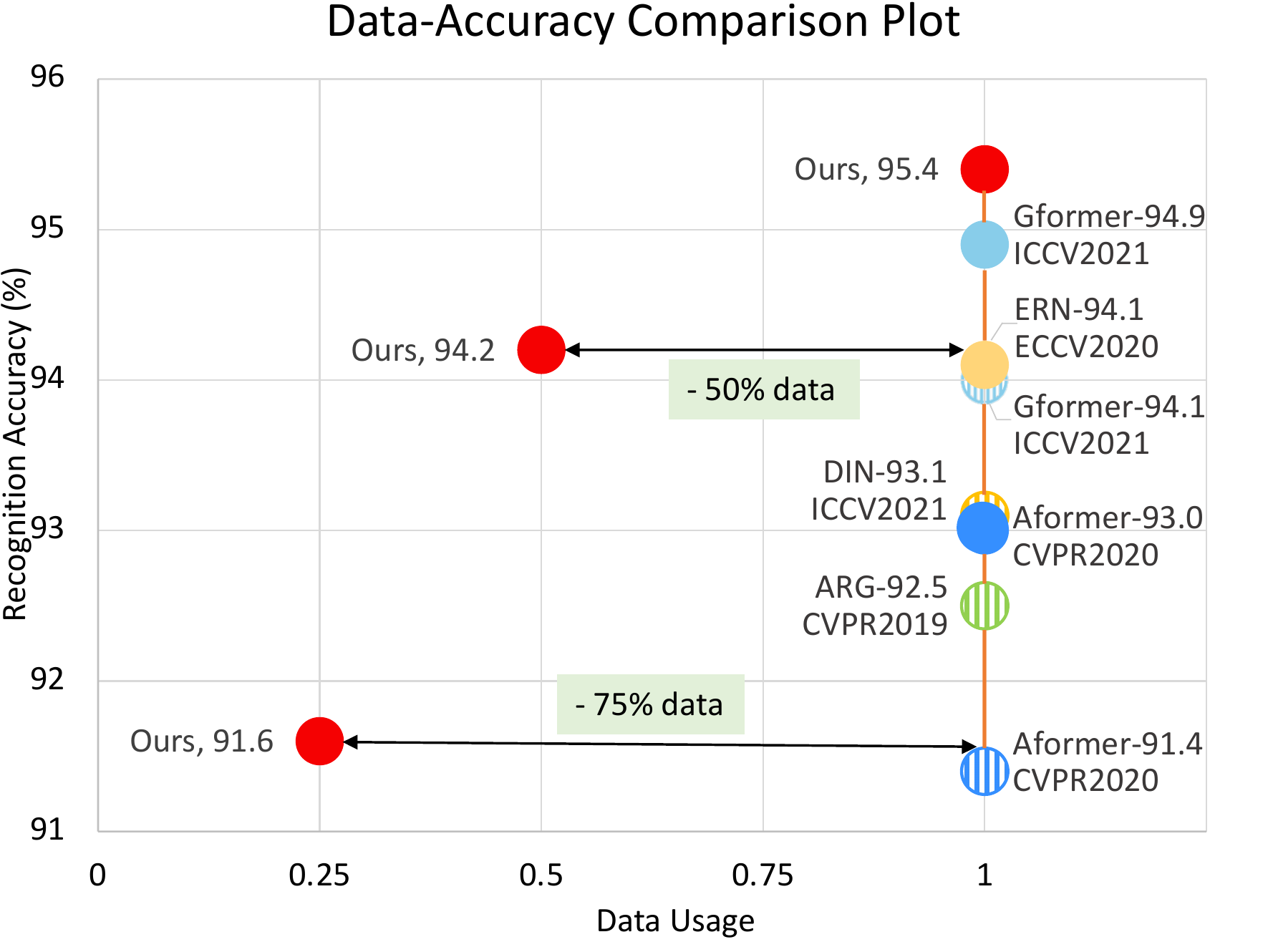}
       \end{center}
          \vspace{-4mm}
          \caption{\textbf{Accuracy comparison with data in different percentage on Volleyball dataset}.
          Our method achieves SOTA performance, 
          and achieves 94.2\% with 50\% data,
          which is competitive to a number of recent approaches \cite{pramono2020empowering,gavrilyuk2020actor,wu2019learning} trained with 100\% data. 
          Solid point means result with additional optical flow input.
          }
         \label{fig:supervision}
    \end{figure}

        Finally,
        we conduct extensive experiments on the widely-used benchmarks to evaluate our designs.
        Our Dual-AI simply achieves state-of-the-art performance on all the fully-annotated datasets,
        such as 
        Volleyball, 
        Collective Activity.
        More interestingly, 
        our Dual-AI with 50\% training data is competitive to a number of recent approaches with 100\% training data in Volleyball as shown in \cref{fig:supervision},
        which clearly demonstrates the generalization power of our Dual-AI.
        Motivated by this,
        we further investigate the challenging setting with limited actor supervision \cite{yan2020social},
        where
        Dual-AI also achieves SOTA results on Weak-Volleyball-M and NBA datasets.
        All these results show that our Dual AI is effective for learning spatiotemporal actor relations in GAR.
    
    \begin{figure*}[t]
       \begin{center}
       \includegraphics[width=1.\linewidth]{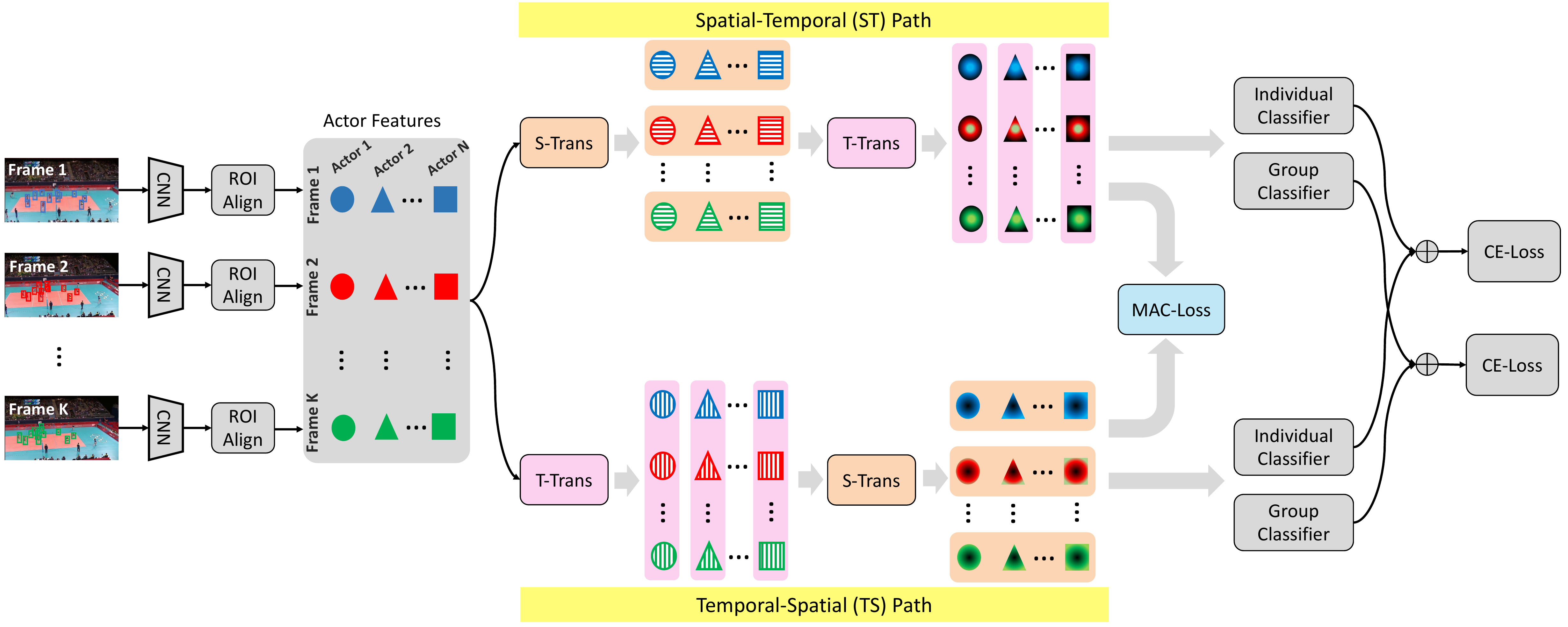}
       \end{center}
      \vspace{-3.5mm}
       \caption{Our Dual-path Actor Interaction (Dual-AI) learning framework,
       where
       S-Trans and T-Trans denote Spatial-Transformer and Temporal-transformer respectively.
       It effectively explores actor evolution in two complementary spatiotemporal views, \ie, ST path and TS path, detailed in \cref{sec:dual-ai}.
       Moreover,
       a Multi-scale Actor Contrastive loss is designed to enable interaction and cooperation of the two paths as in \cref{sec:mac-loss}.
       }
       \label{fig:overall_framework}
      \vspace{-2mm}
    \end{figure*}
    
    \section{Related Work}
    
    \noindent
    \textbf{Group activity recognition} has attracted a large body of work recently due to its wide applications. 
    Early approaches are based on hand-crafted features and typically use probabilistic graphical models \cite{amer2014hirf,amer2013monte,amer2015sum,lan2012social,lan2011discriminative,wang2013bilinear} and AND-OR grammar methods \cite{amer2012cost,shu2015joint}. 
    Recently, 
    methods incorporating convolutional neural networks  \cite{bagautdinov2017social,ibrahim2016hierarchical} and recurrent neural networks \cite{wang2017recurrent,yan2018participation,qi2018stagnet,bagautdinov2017social,deng2016structure,shu2019hierarchical,li2017sbgar,ibrahim2016hierarchical,ibrahim2018hierarchical} have achieve remarkable performance,
    due to the learning of temporal context and high-level information.
    
    More recent group activity recognition methods \cite{wu2019learning,gavrilyuk2020actor,hu2020progressive,yan2020social,ehsanpour2020joint,pramono2020empowering,li2021groupformer,yuan2021spatio} often require the explicit representation of spatiotemporal relations, 
    dedicated to apply attention-based methods to model the individual relations for inferring group activity. 
    \cite{wu2019learning,yuan2021spatio} build relational graphs of the actors and explore the spatial and temporal actor interactions in the same time with graph convolution networks. 
    These methods simulate spatiotemporal interaction of actors in a joint manner.
    Differently,
    \cite{yan2020social} builds separate spatial and temporal relation graphs subsequently to model the actor relations.
    \cite{gavrilyuk2020actor} encodes temporal information with I3D \cite{carreira2017quo} and constructs spatial relation of the actors with a vanilla transformer.
    \cite{li2021groupformer} introduces a cluster attention mechanism for better group informative features with transformers.
    Different from previous approaches,
    we propose to learn the actor interactions in complementary Spatial-Temporal and Temporal-Spatial views and further promote actor interaction learning with a designed self-supervised loss for effective representation learning.
    
    \noindent
    \textbf{Vision Transformer} has gradually become popular for computer vision tasks.
    In image domain, ViT\cite{dosovitskiy2020image} firstly introduces a pure transformer architecture without convolution for image recognition. 
    Following works \cite{li2021ffa, yuan2021tokens,liu2021swin,wang2021pyramid} make remarkable progress on enabling transformer architecture to become a general backbone on various kinds of downstream computer vision tasks. 
    In video domain, 
    many works\cite{han2020mining, arnab2021vivit, li2022uniformer, bertasius2021space,fan2021multiscale,patrick2021keeping} explore spatial and temporal self-attention to learn efficient video representation. TimeSformer\cite{bertasius2021space} investigates the different space and time attention mechanisms to learn spatial-temporal representation efficiently. 
    MViT\cite{fan2021multiscale} utilizes the multi-scale features aggregation to enhance the spatial-temporal representation.  
    Motionformer\cite{patrick2021keeping} presents a trajectory-focused self-attention block, 
    which essentially tracks space-time patches for video transformer. 
    The above transformer architectures are designed for general video classification task.
    It has not been fully explored to tackle the challenging GAR problem with transformers.
    We propose to construct dual spatiotemporal paths with transformers to flexibly learn actor interactions for group activity recognition.
    
    \vspace{-1mm}
    \section{Method}
    \label{sec:method}
    
    To learn complex actor relations in the group activities, 
    we propose a distinct Dual-path Actor Interaction (Dual-AI) framework for GAR. 
    In this section,
    we introduce our Dual-AI in detail.
    First,
    we describe an overview of Dual-AI framework.
    Then,
    we explain how to build the interaction paths,
    with assistance of spatial and temporal transformers.
    Next,
    we introduce a Multi-scale Actor Contrastive Loss (MAC-Loss) to further improve actor consistency between paths.
    Finally,
    we describe the training objectives to optimize our Dual-AI framework.
    
    \subsection{Framework Overview}
    \label{sec:overview}
    As shown in \cref{fig:overall_framework},
    our Dual-AI framework consists of three important steps.
    First,
    we need to extract actor features from backbone.
    Specifically,
    we 
    sample $K$ frames from the input video.
    To make a fair comparison with the previous works in GAR \cite{bagautdinov2017social,li2021groupformer,wu2019learning,yuan2021spatio,yuan2021learning},
    we choose ImageNet-pretrained Inception-v3 \cite{szegedy2016rethinking} as backbone to extract feature of each sampled frame.
    Then,
    we apply RoIAlign \cite{he2017mask} on the frame feature,
    which can generate actor features in this frame from bounding boxes of $N$ actors.
    After that,
    we adopt a fully-connected layer to further encode each actor feature into a $C$ dimensional vector.
    For convenience,
    we denote all the actor vectors as $\mathbf{X} \in \mathbb{R}^{K \times N \times C}$.
    More 
    details can be found in \cref{sec:implement}.
    
    After extracting actor feature vectors,
    we next learn spatiotemporal interactions among these actors in the video.
    Different from the previous approaches \cite{wu2019learning,yuan2021spatio,yan2020social,yan2020higcin,gavrilyuk2020actor},
    we disentangle spatiotemporal modeling into consecutive spatial and temporal interactions in different orders.
    Specifically,
    we design spatial and temporal transformers as basic actor relation modules.
    By flexibly arranging these transformers in two reverse orders,
    we can enhance actor relations with complementary integration of both spatial-temporal (ST) and temporal-spatial (TS) interaction paths.
    Finally,
    we design training losses to optimize our Dual-AI framework.
    In particular,
    we introduce a novel Multi-scale Actor Contrastive Loss (MAC-Loss) between two paths,
    which can effectively improve discriminative power of individual actor representations,
    by actor consistency in all the frame-frame, frame-video, video-video levels.
    Subsequently,
    we integrate actor representations of two paths to recognize individual actions and group activities.

    \subsection{Dual-path Actor Interaction}
    \vspace{-1mm}
    \label{sec:dual-ai}
    To capture complex relations for diversified group activities,
    we propose a novel dual path structure to describe actor interactions.
    To start with, 
    we build basic spatial and temporal actor relation units,
    with assistance of transformers.
    Then,
    we explain how to construct dual paths for spatiotemporal actor interactions.
    
    \subsubsection{Spatial/Temporal Actor Relation Units}
    To understand spatiotemporal actor evolution in videos,
    we first construct basic units to describe spatial and temporal actor relations.
    Since there is no prior knowledge about actor relation,
    we propose to use transformer to model such relation by the powerful self-attention mechanism.

    \textbf{Spatial Actor Transformer.}
    In order to model the spatial relation of the actors in single frame, 
    we design a concise spatial actor transformer ($\mathrm{S\!\!-\!\!Trans}$).
    Specifically,
    we denote $\mathbf{X}^k \in \mathbb{R}^{N\times C}$ as the feature vectors of $N$ actors in the $k$-th frame.
    The spatial relation among these actors are modeled by $\mathbf{\hat{X}}^k={\rm S\!\!-\!\!Trans}(\mathbf{X}^k)$,
    which consists of three modules as follows,
    
    \vspace{-2mm}
    \begin{align}
        \label{eq:s-trans-spe}
        & \mathbf{X}'= {\rm SPE}(\mathbf{X}^k)+\mathbf{X}^k, \\
        \label{eq:s-trans-mhsa}
        & \mathbf{X}'' = {\rm LN}(\mathbf{X}'+{\rm MHSA}(\mathbf{X}')), \\
        \label{eq:s-trans-ffn}
        & \mathbf{\hat{X}}^k = {\rm LN}\big(\mathbf{X}''+{\rm FFN}(\mathbf{X}'')\big). 
    \end{align}
    First,
    we use spatial position encoding (SPE) to add spatial structure information of the actors in the scene,
    as in \cref{eq:s-trans-spe}. 
    We represent spatial position of each actor with center point of its bounding box
    and encode the spatial positions with PE function in \cite{gavrilyuk2020actor,carion2020end}.
    Second,
    we use multi-head self-attention (MHSA) \cite{vaswani2017attention} module to reason the spatial interaction of the actors in the scene,
    as in \cref{eq:s-trans-mhsa}. 
    Finally,
    we use feed-forward network (FFN) \cite{vaswani2017attention} to further improve learning capacity of the spatial actor relation unit,
    as in \cref{eq:s-trans-ffn}.
    
    \textbf{Temporal Actor Transformer.}
    In order to model the temporal evolution of single actor across frames, 
    we design a temporal actor transformer ($\mathrm{T\!\!-\!\!Trans}$) 
    following the way in \cref{eq:s-trans-spe,eq:s-trans-mhsa,eq:s-trans-ffn}. 
    Differently,
    we use the input as the feature vectors of the $n$-th actor across $K$ frames,
    \ie,
    $\mathbf{X}^n \in \mathbb{R}^{K\times C}$.
    In this case,
    the MHSA module can reason the evolution of actor $n$ in different time steps.
    Moreover,
    to add temporal sequence information of actor $n$,
    temporal position encoding (TPE) is used instead of SPE,
    which encodes frame index $\{1,...,K\}$ with PE function in \cite{vaswani2017attention}.
    Finally,
    we can get actor features enhanced by temporal interactions, as $\mathbf{\hat{X}}^n={\rm T\!\!-\!\!Trans}(\mathbf{X}^n)$.

    \subsubsection{Dual Spatiotemporal Paths of Actor Interaction}
    \label{sec:compl_patt}
    
    Once the spatial and temporal relations of actors are built,
    we can further integrate them to construct spatiotemporal representation of the actor evolution.
    As discussed in \cref{sec:intro},
    the single order of space and time is insufficient to understand the complex actor interactions,
    leading to the failure of inferring group activities.
    Thus,
    we propose a dual spatiotemporal paths framework for GAR to capture the complex interaction of the actors.
    
    It consists of two complementary spatiotemporal modeling patterns for actor evolution, \ie, Spatial-Temporal (ST) and Temporal-Spatial (TS),
    by switching the order of space and time as: 
    \begin{align}
        \label{eq:ST}
        \mathbf{X}_{\mathrm{ST}}&={\rm T\!\!-\!\!Trans}(\mathbf{X}+{\rm MLP}({\rm S\!\!-\!\!Trans}(\mathbf{X}))) \\
        \label{eq:TS}
        \mathbf{X}_{\mathrm{TS}}&={\rm S\!\!-\!\!Trans}(\mathbf{X}+{\rm MLP}({\rm T\!\!-\!\!Trans}(\mathbf{X}))),
    \vspace{-2mm}
    \end{align}
    where we adopt a residual structure to enhance the actor representation. 
    MLP with parameters in shape $C\times C$ is used to add non-linearity. 
    By reshaping the frame and actor dimension as batch dimension,
    $\mathrm{S\!\!-\!\!Trans}$ and $\mathrm{T\!\!-\!\!Trans}$ reason about spatial and temporal actor interaction respectively.
    
    By stacking spatial and temporal transformers in different orders, 
    the actor representation is reweighted and aggregated according to different spatiotemporal context.
    ST path first reasons about the interaction of different actors in the scene of each frame.
    Then, 
    the temporal evolution is modeled to reweight the built actor interaction across different frames. 
    As such, 
    ST path is skilled at recognizing activities with distinct spatial arrangement,
    such as \textit{set} in volleyball games.
    This activity requires the player to move to a new position and set the ball, 
    usually accompanied by other players moving or jumping for fake spiking.
    Complementarily,
    TS path reasons about the actor evolution, 
    in the opposite order of ST path.
    It considers temporal dynamics of each actor in the first place, 
    and then reasons about spatial actor interaction to understand the scene.
    Hence,
    it is skilled at recognizing activities with distinct actor evolution patterns,
    such as \textit{spike} in volleyball games,
    which requires hitter to jump and quickly hit the ball.
     
    Subsequently,
    to fully take advantage of such complementary characteristic,
    we feed the representation of actors from ST and TS paths to generate individual actions and group activity predictions,
    and  
    fuse them as final predictions of dual spatiotemporal paths.

    \subsection{Multi-scale Actor Contrastive Learning}
    \label{sec:mac-loss}
    
    The actor representation is reweighted and aggregated by dual spatiotemporal paths,
    however,
    the modeling process is independent.
    To promote cooperation of these two complementary paths,
    we design a self-supervised Multi-scale Actor Contrastive loss (MAC-loss). 
    As dual spatiotemporal paths model evolution of each actor in different patterns,
    we define a pretext task of actor consistency. 
    Specifically,
    we design such constraints in multiple scales of frame and video levels.

    \begin{figure}
      \begin{center}
      \includegraphics[width=1.\linewidth]{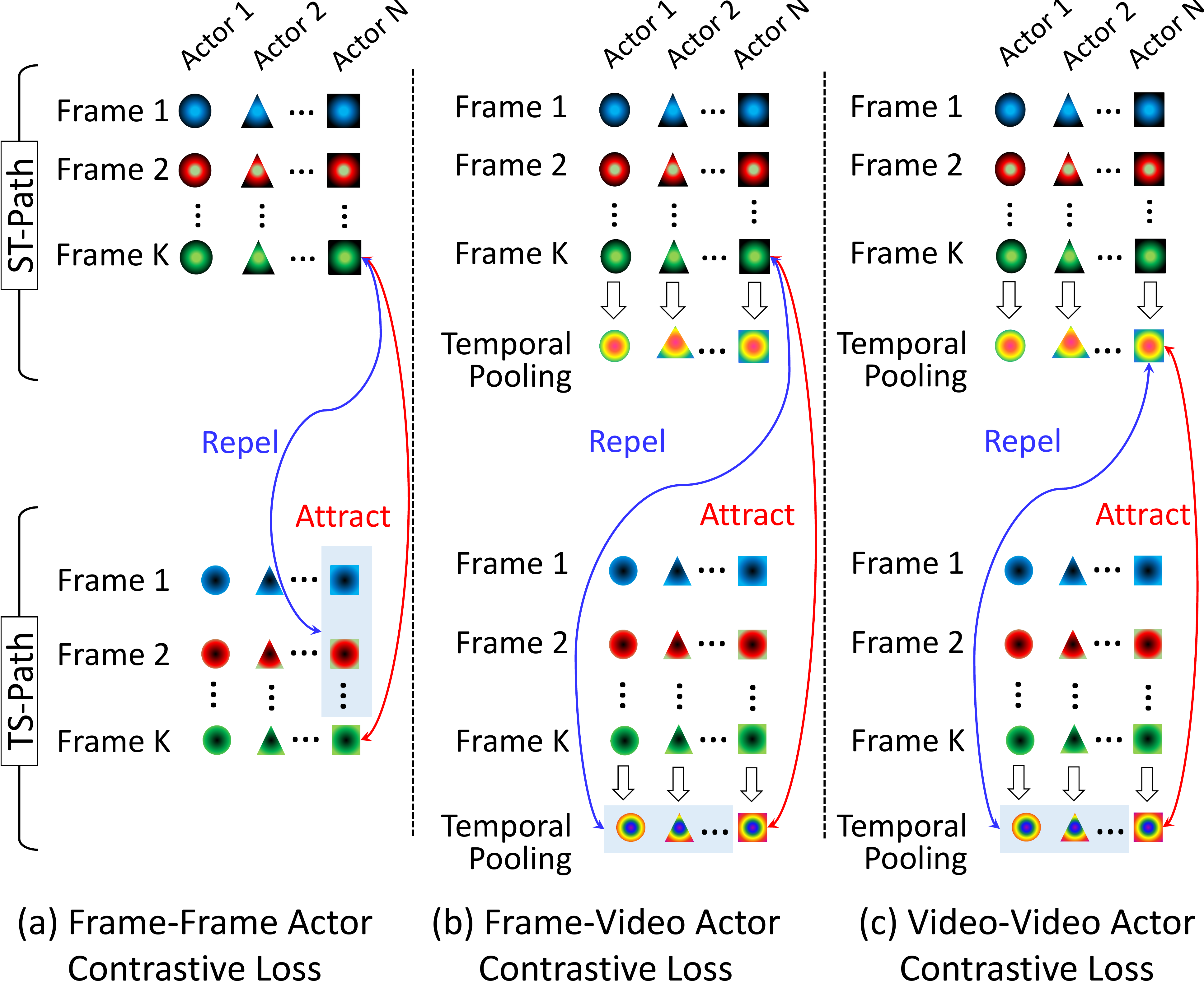}
      \end{center}
          \caption{Illustration of MAC-loss for Actor N.
          It consists of three levels, \ie, frame-frame, frame-video and video-video.
          The blue block means the source of negative pairs.
          For simplicity,
          we only show the constraints from ST path to TS path.
          It is similar for the constraints from TS path to ST path.
          }
         \label{fig:consistency}
    \end{figure}
    
    \vspace{+1mm}
    \textbf{Frame-Frame Actor Contrastive Loss.}
    The frame representation of the actor in one path should be similar with its corresponding frame representation in the other path,
    while different from other frame representation of this actor in the path.
    As shown in \cref{fig:consistency} (a),
    taking actor $n$ in ST path as an example,
    we attract frame representation in $k$-th frame ($\mathbf{X}_{\mathrm{ST}}^{n,k}$) to its corresponding representation from TS path ($\mathbf{X}_{\mathrm{TS}}^{n,k}$).
    Meanwhile,
    we repel the representation of actor $n$ in other frames from TS path ($\mathbf{X}_{\mathrm{TS}}^{n,t}$, where $t\!\neq \!k$), 
    \begin{equation}
        \label{eq:consis_ff}
        \mathcal{L}_{f\!f}(\mathbf{X}^{n,k}_{\mathrm{ST}},\mathbf{X}^{n,k}_{\mathrm{TS}})=-\log \frac{h(\mathbf{X}^{n,k}_{\mathrm{ST}}, \mathbf{X}^{n,k}_{\mathrm{TS}})}{\sum_{t=1}^{K}h(\mathbf{X}^{n,k}_{\mathrm{ST}}, \mathbf{X}^{n,t}_{\mathrm{TS}})},
    \end{equation}
    where 
    $h(\mathbf{u},\mathbf{v}) = \exp(\frac{\mathbf{u}^\top \mathbf{v}}{||\mathbf{u}||_2||\mathbf{v||_2}})$ is the exponential of cosine similarity measure.
    Vice versa,
    the loss for actor $n$ in TS path can be obtained by $\mathcal{L}_{f\!f}(\mathbf{X}^{n,k}_{\mathrm{TS}},\mathbf{X}^{n,k}_{\mathrm{ST}})$.
    
    \vspace{+1mm}
    \textbf{Frame-Video Actor Contrastive Loss.}
    The frame representation of the actor in one path should be consistent with its video representation in the other path,
    while different from video representation of other actors in the path.
    As shown in \cref{fig:consistency} (b),
    taking actor $n$ in ST path as an example,
    we attract its frame representation $\mathbf{X}_{\mathrm{ST}}^{n,k}$ to its video representation $\mathbf{\tilde{X}}^{n}_{\mathrm{TS}}$ from TS path,
    which is obtained by pooling frame representation $\mathbf{X}_{\mathrm{TS}}^{n,{1:K}}$.
    Meanwhile,
    we repel the video representation of other actors in the minibatch from TS path ($\mathbf{\tilde{X}}^{i}_{\mathrm{TS}}$, where $i\!\neq \!n$), 
    \vspace{-1mm}
    \begin{equation}
        \label{eq:consis_lg}\mathcal{L}_{f\!v}(\mathbf{X}^{n,k}_{\mathrm{ST}},\!\mathbf{\tilde{X}}^{n}_{\mathrm{TS}})=-\log \frac{h(\mathbf{X}^{n,k}_{\mathrm{ST}}, \mathbf{\tilde{X}}^{n}_{\mathrm{TS}})}{\sum_{i=1}^{B\times N}h(\mathbf{X}^{n,k}_{\mathrm{ST}}, \mathbf{\tilde{X}}^{i}_{\mathrm{TS}})},
    \vspace{-1mm}
    \end{equation}
    where $B$ denotes the minibatch size. 
    Vice versa,
    the loss for actor $n$ in TS path can be obtained by $\mathcal{L}_{f\!v}(\mathbf{X}^{n,k}_{\mathrm{TS}},\mathbf{\tilde{X}}^{n}_{\mathrm{ST}})$.
    
    \vspace{+1mm}
    \textbf{Video-Video Actor Contrastive Loss.}
    Furthermore,
    we constrain the consistency of video representation of each actor across dual paths,
    as shown in \cref{fig:consistency} (c).
    We achieve this by minimizing cosine similarity measure $\mathcal{L}_{v\!v}$ of corresponding video representation $(\mathbf{\tilde{X}}^{n}_{\mathrm{TS}},\mathbf{\tilde{X}}^{n}_{\mathrm{ST}})$.
    Our proposed MAC-loss is then formed as
    \begin{equation}
        \label{eq:f_contras}
        \mathcal{L}_{M\!AC} = \lambda_{f\!f}\mathcal{L}_{f\!f}+\lambda_{f\!v}\mathcal{L}_{f\!v}+\lambda_{v\!v}\mathcal{L}_{v\!v},
    \end{equation}
    where $\lambda_{\{\cdotp\}}$ denote weights for the different components.
    

    \subsection{Training objectives}
    \label{sec:train_obj}
    Our network can be trained in an end-to-end manner to simultaneously predict individual actions of each actor and group activity.
    Combining with standard cross-entropy loss, the final loss for recognition is formed as
    \begin{equation}
        \small
        \label{eq:recog_loss}
        \mathcal{L}_{cls}\!\! = \!\!\mathcal{L}_{CE}(\frac{\hat{y}^G_{\mathrm{ts}}\!+\!\hat{y}^G_{\mathrm{st}}+\hat{y}^G_{scene}}{3}, y^G) + \lambda \mathcal{L}_{CE}(\frac{\hat{y}^I_{\mathrm{ts}}\!+\!\hat{y}^I_{\mathrm{st}}}{2}, y^I),
    \end{equation}
    where $\hat{y}_{\{ts,st\}}^I$ and $\hat{y}_{\{ts,st\}}^G$ denote individual action and group activity predictions from TS and ST paths. 
    $y^I$ and $y^G$ represent the ground truth labels for the target individual actions and group activity. 
    $\hat{y}^G_{scene}$ denotes the scene prediction produced by separate group activity classifier,
    using features directly from backbone.
    $\lambda$ is the hyper-parameter to balance the two items.
    Finally,
    we combine all the losses to train our Dual-AI framework,
    \begin{equation}
    \mathcal{L} = \mathcal{L}_{cls} + \mathcal{L}_{M\!AC}.
    \end{equation}
    
    During inference,
    we infer the individual actions and group activity by averaging the predictions from the dual spatiotemporal paths.
    
    %
    %
    
    \section{Experiments}
    \subsection{Dataset}
    \label{sec:dataset}
    \noindent
    \textbf{Volleyball Dataset.} 
    This dataset \cite{ibrahim2016hierarchical} consists of 4,830 labeled clips (3493/1337 for training/testing) from 55 volleyball games. 
    Each clip is annotated with one of 8 group activity classes.
    Middle frame of each clip is annotated with 9 individual action labels
    and their bounding boxes.
    
    \noindent
    \textbf{Collective Activity Dataset.}
    This dataset \cite{choi2009they} contains 44 short videos with every ten frames annotated with individual action labels
    and their bounding boxes. 
    The group activity class of each clip is determined by the largest number of the individual action classes. 
    We follow \cite{yan2020higcin,yan2018participation,yuan2021spatio} to merge the \textit{crossing} and \textit{walking} into \textit{moving}.
    
    \noindent
    \textbf{Weak-Volleyball-M Dataset.} 
    This dataset \cite{yan2020social} is adapted from Volleyball dataset while merging \textit{pass} and \textit{set} categories to have total 6 group activity classes, and discarding all individual annotations (including individual action labels and bounding boxes) for weakly supervised GAR.
    
    \noindent
    \textbf{NBA Dataset.} 
    This dataset \cite{yan2020social} contains 9,172 annotated clips (7624/1548 for training and testing) from 181 NBA game videos,
    each of which belongs to one of the 9 group activities. 
    No individual annotations, 
    such as individual action labels and bounding boxes, 
    are provided.
    
    \begin{table}[t]
        \centering
        \footnotesize
        \setlength{\tabcolsep}{0.7mm}{
        \begin{tabular}{lclccc}
        \hline
        Method         & Backbone           & \tabincell{c}{Data \\ Ratio} & \tabincell{c}{Optical \\ Flow}         & \tabincell{c}{Individual \\ Action}   & \tabincell{c}{Group \\ Activity} \\
        \hline
         HDTM\cite{ibrahim2016hierarchical}   &AlexNet              &   100\%         &                &-                   & 81.9      \\
         CERN\cite{shu2017cern}  & VGG16              &   100\%         &                   & -                   & 83.3      \\
         StageNet\cite{qi2018stagnet}& VGG16          &   100\%         &                   & -                 & 89.3     \\
        HRN\cite{ibrahim2018hierarchical} & VGG19              &   100\%         &                   & -                 & 89.5      \\
        SSU\cite{bagautdinov2017social} & Inception-v3       &   100\%         &                   & 81.8              & 90.6      \\
        AFormer\cite{gavrilyuk2020actor} & I3D       &   100\%         &                   & -              & 91.4      \\
        ARG\cite{wu2019learning}  & Inception-v3          &   100\%         &                   & 83.0               & 92.5      \\
        TCE+STBiP \cite{yuan2021learning} & Inception-v3          &   100\%         &                   & -               & 93.3      \\
        DIN \cite{yuan2021spatio} &   ResNet-18      &   100\%         &                   & -             & 93.1 \\
        GFormer\cite{li2021groupformer} &   Inception-v3       &   100\%         &                   & 83.7              & 94.1 \\
        \hline
        \multirow{2}[1]{*}{Ours}    & Inception-v3          &   25\%          &                   &    82.1   & 89.7     \\ 
                 & Inception-v3          &   50\%          &                   &    83.0   & 92.7     \\
                 & Inception-v3          &   100\%         &                   &    \textbf{84.4}      & \textbf{94.4}     \\
        \hline
        SBGAR\cite{li2017sbgar}  & Inception-v3                      &   100\%         &\checkmark                   &  -                 & 66.9     \\
        CRM\cite{azar2019convolutional}  & I3D                &   100\%        & \checkmark        &          -        & 93.0    \\
        Aformer\cite{gavrilyuk2020actor}  & I3D                &   100\%        & \checkmark                  &        83.7           & 93.0      \\
        JLSG\cite{ehsanpour2020joint}   & I3D        &   100\%        & \checkmark                  &        83.3           & 93.1      \\
        ERN\cite{pramono2020empowering}  & R50-FPN+I3D        &   100\%        & \checkmark                  &      81.9             & 94.1      \\
        GFormer\cite{li2021groupformer} &   I3D       &   100\%         & \checkmark                  & 84.0             & 94.9 \\
        \hline
        \multirow{3}[1]{*}{Ours}    & Inception-v3          &   25\%         & \checkmark                  &    83.0           & 91.6     \\
                 & Inception-v3          &   50\%         & \checkmark                  &    84.0           & 94.2     \\
                 & Inception-v3          &   100\%           & \checkmark                  &   \textbf{85.3}        & \textbf{95.4}  \\
        \hline
        \end{tabular}}
        \caption{Comparison with state-of-the-art methods on \textbf{Volleyball dataset} in term of Acc.\%.
        }
        \label{tab:volleyball_full}
    \end{table}
    \begin{table}[t]
    \footnotesize
    \centering
    \setlength{\tabcolsep}{3.5mm}{
    \begin{tabular}{ccc}
        \hline
        Method & Backbone & MPCA \\
        \hline
        HDTM\cite{ibrahim2016hierarchical}  & AlexNet & 89.7  \\
        PCTDM\cite{yan2018participation} & AlexNet & 92.2  \\
        CERN-2\cite{shu2017cern} & VGG-16 & 88.3  \\
        Recurrent\cite{Wang_2017_CVPR} & VGG-16 & 89.4 \\
        stagNet\cite{qi2018stagnet} & VGG-16 & 89.1  \\
        SPA+KD\cite{tang2018mining} & VGG-16 & 92.5  \\
        PRL\cite{hu2020progressive}   & VGG-16 & 93.8  \\
        CRM\cite{azar2019convolutional} & I3D & 94.2  \\
        ARG\cite{wu2019learning}   & ResNet-18 & 92.3  \\
        HiGCIN\cite{yan2020higcin} & ResNet-18 & 93.0  \\
        DIN\cite{yuan2021spatio} & ResNet-18 & 95.3 \\
        TCE+STBiP\cite{yuan2021learning} & Inception-v3 & 95.1  \\
        \hline
        \multirow{2}[1]{*}{Ours}     & ResNet-18 & \textbf{96.0}  \\
                                     & Inception-v3 & \textbf{96.5} \\
        \hline
        \end{tabular}%
    }
        \caption{Comparisons with previous state-of-the-art methods on \textbf{Collective Activity datatset}.}
        \vspace{-4mm}
        \label{tab:collective}%
    \end{table}
    
    \begin{table}[t]
    \parbox{1.\linewidth}{
    \centering
    \footnotesize
    \setlength{\tabcolsep}{0.7mm}{
    \begin{tabular}{llccc}
    \hline
        Method & Backbone  & \tabincell{c}{Mod-\\ ality} & \tabincell{c}{NBA \\ Acc./Mean Acc.}   & \tabincell{c}{Weak Vlb. \\ -M Acc.} \\
        \hline
        TSN*\cite{wang2016temporal}      & Incep-v1   &  RGB        & -- / 37.8 &--  \\
        
        I3D*\cite{carreira2017quo}     &  I3D    &  RGB        & -- / 32.7 &--  \\
        Nlocal*\cite{wang2018non} &  I3D-NLN   & RGB      & -- / 32.3 &--  \\
        ARG*\cite{wu2019learning}   & Incep-v3 & RGB      &-- / --  & 90.7  \\
        SAM\cite{yan2020social}  & Res-18 & RGB      &-- / --  & 93.1 \\
        SAM\cite{yan2020social}           & Incep-v3 & RGB       & 49.1 / 47.5 & 94.0 \\
        \hline
        \multirow{3}[1]{*}{Ours}   & Incep-v3 & RGB        & 51.5 / 44.8 & 95.8  \\
               & Incep-v3 & Flow       & 56.8 / 49.1 & 96.1   \\
                & Incep-v3 & Fusion     & \textbf{58.1} / \textbf{50.2} & \textbf{96.5}   \\
    \hline
    \end{tabular}}
    \vspace{+2mm}
    \caption{Comparision with state-of-the-art methods on \textbf{NBA 
    and Weak-Volleyball-M dataset} 
    following metrics adopted in \cite{yan2020social}. 
    * means the results are from \cite{yan2020social}.}
    \label{tab:volleyball_weak}
    }
    
    \vspace{+6mm}
    \parbox{1.\linewidth}{
    \footnotesize
    \centering
        \begin{tabular}{lccccc}
        \hline
        Method           & 5\% &  10\%  & 25\%  & 50\%   & 100\% \\
        \hline
        PCTDM\cite{yan2018participation}  & 53.6 & 67.4 & 81.5 & 88.5 & 90.3    \\
        AFormer\cite{gavrilyuk2020actor}  &  54.8   &  67.7    &  84.2   & 88.0    & 90.0    \\
        HiGCIN\cite{yan2020higcin}        & 35.5  & 55.5    &  71.2 & 79.7    & 91.4    \\
        ERN\cite{pramono2020empowering} &  41.2    &  52.5   &73.1    & 75.4      & 90.7    \\
        ARG\cite{wu2019learning}          & 69.4 & 80.2 & 87.9 & 90.1 & 92.3    \\
        DIN\cite{yuan2021spatio}          & 58.3 & 71.7  & 84.1 & 89.9  & 93.1    \\
        \hline
        Ours                             & \textbf{76.2}  & \textbf{85.5} & \textbf{89.7} & \textbf{92.7} & \textbf{94.4}    \\
        \hline
        \end{tabular}
        \caption{Comparison with state-of-the-art methods trained with Volleyball dataset of different data ratios in term of group activity recognition Acc.\%.}
        \label{tab:volleyball_limited}
    }
    \end{table}
    
    \subsection{Implementation Details}
    \label{sec:implement}
    We select the Inception-v3 model as our CNN backbone, 
    following widely used settings \cite{bagautdinov2017social,li2021groupformer,wu2019learning,yuan2021spatio,yuan2021learning} in GAR.
    We also use ResNet-18 model as backbone for Collective Activity Dataset, following widely used settings \cite{yan2020higcin,yuan2021spatio}.
    We apply the ROI-Align with crop size $5\times 5$ and a linear embedding to get actor features with dimension $C=1024$.
    Each Spatial or Temporal transformer has one attention layer with 256 embedding dimension. 
    The $\lambda_{f\!f}, \lambda_{f\!v}, \lambda_{v\!v}$ in MAC-Loss are all set 1.
    More details for $K$ and $N$ can be found in supplementary material.
    
    
        
\subsection{SOTA Comparison}
\label{sec:sota}
 \textbf{Full Setting.}
 This setting allows us to train our model with all data fully annotated with group activities and individual annotations. 
 We compare our method with the state-of-the-art approaches on Volleyball and Collective Activity dataset.
 As shown in \cref{tab:volleyball_full},
 our approach (94.4\%) with only RGB frames and Inception backbone has already outperformed other SOTA methods with computationally high backbones (I3D, FPN) and additional optical flow input. 
 Furthermore, 
 equipped with RGB and optical flow late fusion, 
 our method can improve the SOTA result by a large margin to 95.4\%. 
Remarkably, 
even with only 50\% data, 
our method still surpasses the vast majority of the SOTA methods with 100\% data, 
\eg, 
Ours (50\%) vs. SARF (100\%): 94.2 vs. 93.1.
 As shown in \cref{tab:collective},
 our approach also achieves state-of-the-art performance on Collective Activity dataset. 
 These results demonstrate the effectiveness of our method. 
 
\textbf{Weakly Supervised Setting.}
 Under this setting we use all raw data and group activity annotations, 
 without any individual annotations.
 We follow the \cite{yan2020social} to report results on Weak-Volleyball-M dataset and NBA dataset.
As shown in \cref{tab:volleyball_weak}, 
our method surpasses all the existing methods by a good margin, 
establishing new state-of-the-art results. 
Specifically, 
our approach improves the previous SOTA \cite{yan2020social} by 2.5\% on Weak-Volleyball-M and by 9\% on NBA dataset in term of Acc.\%. 
It indicates that our Dual-AI framework can enhance the learning ability of the model to obtain robust representation and achieve promising performance in the case individual annotations missing.

\textbf{Limited Data Setting.}
 In this setting, 
 we train our method with random sampled 
 data in different ratios to show the generalization power of our method.
To compare the results under this setting,
we implement a number of previous SOTA methods that have the officially-published codes available.
 As shown in \cref{tab:volleyball_limited}, 
 our method surpasses previous SOTA methods in all data ratios.  
 Moreover,
 with the available training data decreasing, 
 the performance of our method remains promising and the gain against other methods gets enlarged,
 which demonstrates the robustness of our method.

\begin{table}[t]
    \centering
    \footnotesize
    \setlength{\tabcolsep}{1.5mm}{
    \begin{tabular}{lccc}
    \hline
    Dual-Path & \tabincell{c}{Weak \\ Volleyball-M} & \tabincell{c}{Limited \\Volleyball} & \tabincell{c}{Full \\Volleyball}\\
    \hline
     S-S           & 88.9 & 88.4 & 91.2  \\
     T-T           & 91.6 & 87.9 & 90.9  \\
     S-T            & 93.0 & 89.3& 92.2 \\
     T-S            & 92.6 & 89.5& 92.1 \\
     ST-TS Fusion  & \textbf{94.2} & \textbf{90.8} & \textbf{93.3} \\
    \hline
    \end{tabular}
    }
    \caption{
    Effectiveness of our Dual Path Actor Interaction. 
    }
    \label{tab:ablation1}
\end{table}
\begin{table}
    \centering
    \footnotesize
    \setlength{\tabcolsep}{3.5mm}{
    \begin{tabular}{ccc|cc}
    \hline
    \multicolumn{3}{c|}{Components of MAC-loss} & \multicolumn{2}{c}{Data Ratio} \\
    \hline
     F-F & F-V &V-V & 50\%   & 100 \% \\
    \hline
                 &             &            &90.8     & 93.3  \\
    \checkmark   &             &            &91.2     & 93.5\\
                 &  \checkmark &            &91.0     & 93.3\\
                 &             &\checkmark  &91.6     & 93.6\\
    \checkmark   &  \checkmark &\checkmark  &\textbf{92.1}     & \textbf{\textbf{94.0}}\\
    
    \hline
    \end{tabular}
    }
    \caption{Effectiveness of our MAC-loss. 
    Different components are ablated on Volleyball dataset in term of Acc.\%.}
    \label{tab:ablation2}
\end{table}
    
\begin{figure*}[tb]
\begin{center}
\includegraphics[width=0.96\linewidth]{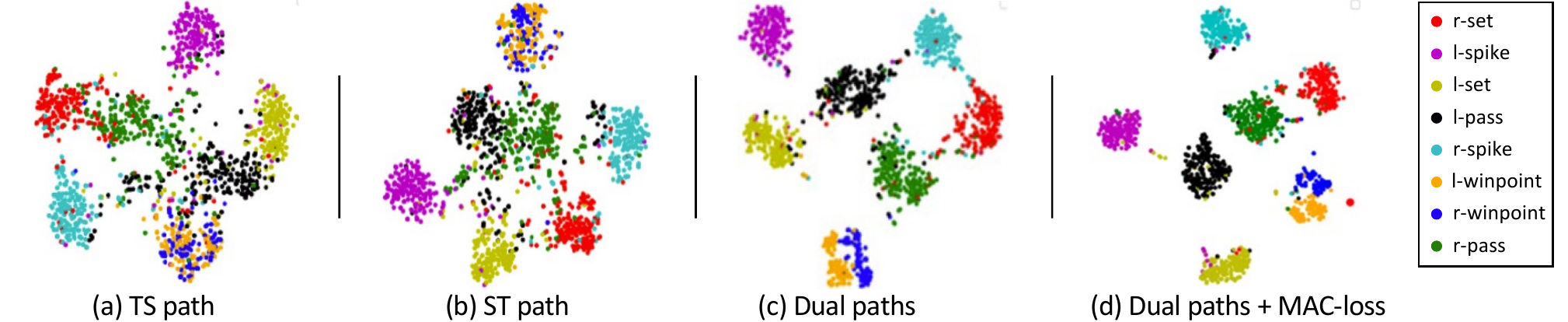}
\end{center}
  \vspace{-5mm}
  \caption{t-SNE \cite{van2008visualizing} visualization of video representation on the Volleyball dataset learned by different variants of our Dual-AI model: 
  ST path only, 
  TS path only, 
  Dual spatiotemporal paths, 
  and final Dual-AI model.
  }
 \label{fig:vis_tsne}
\end{figure*}

\begin{figure*}
\begin{center}
\includegraphics[width=1.\linewidth]{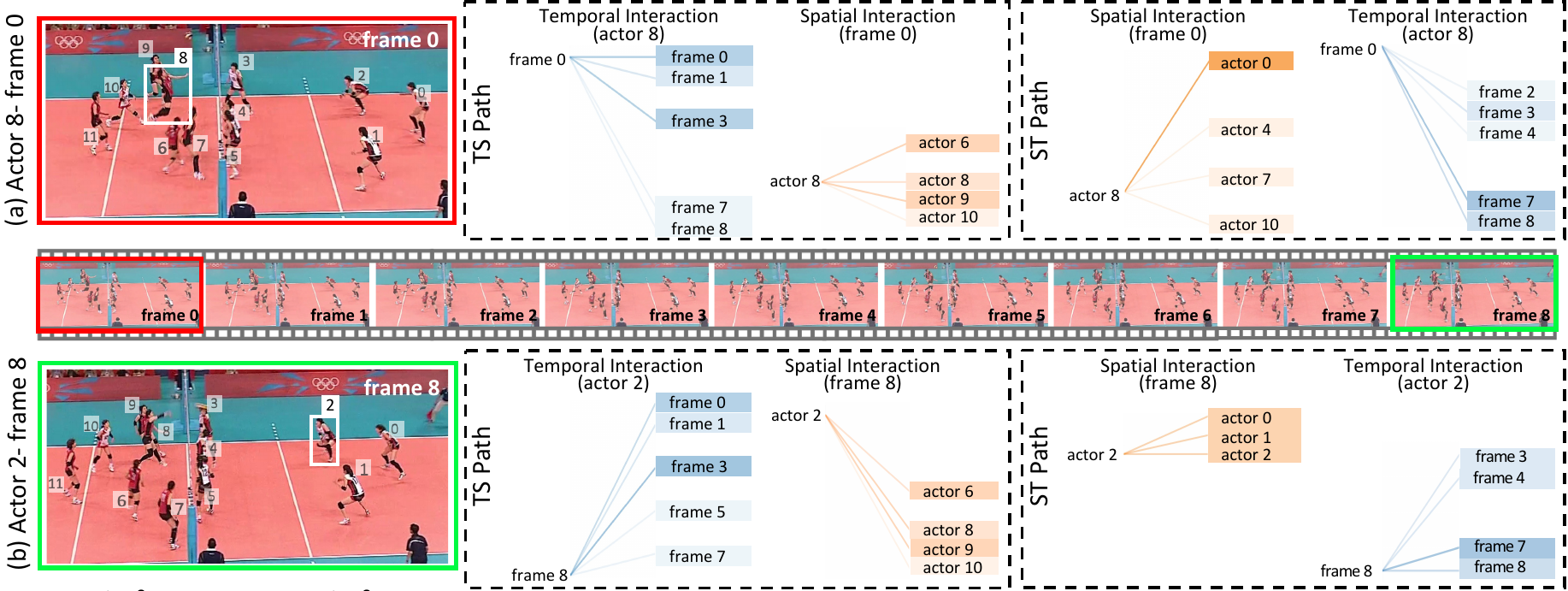}
\end{center}
  \vspace{-4mm}
  \caption{Actor interaction visualization for \textit{l-spike} activity with connected lines.
  Brighter color indicates stronger relation.
  (a) For actor 8 in frame 0, we visualize the temporal interaction with same actors in different frames for ST and TS paths; 
  similarly,
  we visualize the spatial interaction with different actors in frame 0.
  (b) We visualize the actor interaction for actor 2 in frame 8 in the same way.
  }
 \label{fig:vis_attention}
\end{figure*}
 
 \subsection{Ablation Study }
 
 \textbf{Dual Spatial Temporal Paths.}
 To validate the effectiveness of our Dual Spatiotemporal Paths, 
 we investigate six settings. 
 Particularly, we experiment with 50\% data for limited Volleyball. 
 In addition to T-S and S-T introduced in Section \cref{sec:dual-ai},
 other two paths, \ie, S-S and T-T are introduced to validate in a broader range.
 S-S/T-T means that features go through two successive Spatial/Temporal-Transformer, respectively. 
 As shown in \cref{tab:ablation1}, 
 our Dual Paths achieves the best result under different setting.
 The reason is that,
 dual-path TS and ST are good at inferring different group activities and the learned representation from ST and TS can complement each other, leading to a better performance. 
 This demonstrates that our dual path ST-TS is a preferable way to comprehensively leverage both paths to generate robust spatiotemporal contexts for boosting group activity recognition.
 
\textbf{Multi-scale Actor Contrastive Loss.}
 We explore the performance of our network with different components of MAC loss.
 As shown in \cref{tab:ablation2}, 
 with different component of consistent loss (frame-frame, frame-video, video-video), 
 our network consistently outperforms w/o consistent loss. 
 By utilizing all components of MAC-loss, 
 our network can achieve the best results. 
 Note that,
 given less available training data, 
 the loss can help network get a larger accuracy improvement. 
 It demonstrates that the MAC-loss can enable cooperation of the dual complementary modeling process,
 thereby enhancing the learned representation from ST and TS paths, 
 especially with limited available data.

\begin{table}[t]
\parbox{1.\linewidth}{
    \centering
    \footnotesize
    \setlength{\tabcolsep}{4.5mm}{
    \begin{tabular}{ccc}
    \hline
     \multirow{2}{*}{Scene Fusion} & \multicolumn{2}{c}{Data Ratio}\\ \cline{2-3}
                                  & 50\%                & 100\%   \\ 
    \hline
    w/o       & 92.1 & 94.0\\
    Early     & 92.0 & 93.9\\   
    Middle    & 92.2 & 94.0\\
    Late      & \textbf{92.7} & \textbf{94.4}\\
    
    \hline
    \end{tabular}
    }
    \caption{Effectiveness of scene information. 
    }
    \vspace{-2mm}
    \label{tab:ablation3}

}
\end{table}

\textbf{Scene Information.}
 We investigate the effectiveness of scene information,
 by exploring the way to fuse scene context in a early, middle and late fusion manner.
 As shown in \cref{tab:ablation3}, 
 late scene context fusion is the best choice.  
 Regardless of the available data ratio, 
 the scene information can improve the performance by around 0.6 in term of Acc.\%. 
 This is because that scene information can provide global-level context, which can supplement the actor-level relation modeling and is crucial to GAR.

    \subsection{Visualization}

    \vspace{-2mm}
    \textbf{Group Feature Visualization.}
    \cref{fig:vis_tsne} shows the t-SNE \cite{van2008visualizing} visualization of the learned representation. 
    We project video representation extracted from Volleyball validation dataset to 2-D dimension using t-SNE. 
    We can see that learned representation from Dual Path transformer (c) can be grouped better than single Temporal-Spatial path (a) and Spatial-Temporal path (b). 
    Furthermore,  
    equipped with MAC-loss, 
    our Dual-AI network (d) is able to differentiate group representations much better. 
    These results demonstrate the effectiveness of our Dual-AI framework. 
    
    \textbf{Spatial/Temporal Actor Attention Visualization.}
    We visualize the actor interaction of \textit{l-spike} activity in \cref{fig:vis_attention}. 
    The attention weight between actors is represented by connected lines,
    and the brightness of the lines represents the scale of the attention weight. 
    Orange and Blue lines correspond to the Spatial and Temporal interaction, 
    respectively. 
    As shown by spatial interaction in \cref{fig:vis_attention} (a), 
    the spiking player (actor 8) is more related with accompanying players in TS path, who are ``moving" (actor 6 and 10) and ``standing" (actor 9).
    Differently,
    in ST path,
    actor 8 has wider connections with accompanying players (\eg, actor 7 and actor 10) and defending players (\eg, actor 0 and actor 4).
    Similarly,
    as shown by spatial interaction in \cref{fig:vis_attention} (b),
    the actor 2 is related to different accompanying and defending players in TS path and ST path respectively,
    showing complementary patterns.
    As for temporal interaction in both (a) and (b),
    the anchor actor is more related with early frames (frame 0 and frame 3) in TS path,
    while more related with late frames (frame 7 and frame 8) in ST path, showing highly complementary patterns. 
    
    \vspace{-1mm}
    \section{Conclusion}
    In this work,
    we develop a Dual-AI framework to flexibly learn actor interactions in Spatial-Temporal and Temporal-Spatial views.
    Furthermore,
    we design a distinct MAC-loss to enable cooperation of dual paths for effective actor interaction learning.
    We conduct experiments on three datasets and establish new state-of-the-art results under different data settings. 
    Particularly,
    our method with 50\% data surpasses a number of recent methods trained with 100\% data.
    The comprehensive ablation experiments and visualization results show that our method is able to learn actor interaction in a complementary way.
    
    \section*{Acknowledgement}
    This work is partially supported by the National Natural Science Foundation of China (61876176, U1813218), the Joint Lab of CAS-HK, Guangdong NSF Project (No.2020B1515120085), the Shenzhen Research Program (RCJC20200714114557087), the Shanghai Committee of Science and Technology, China (Grant No.21DZ1100100). This work is also partially supported by Australian Research Council Discovery Early Career Award (DE190100626).
    
    {\footnotesize
    \bibliographystyle{ieee_fullname}
    \bibliography{egbib}
    }
    
    \clearpage

\twocolumn[{%
\renewcommand\twocolumn[1][]{#1}
\maketitle
\begin{center}\
\renewcommand\thefigure{C.1}
  \centering
  \captionsetup{type=figure}
  \includegraphics[width=\linewidth]{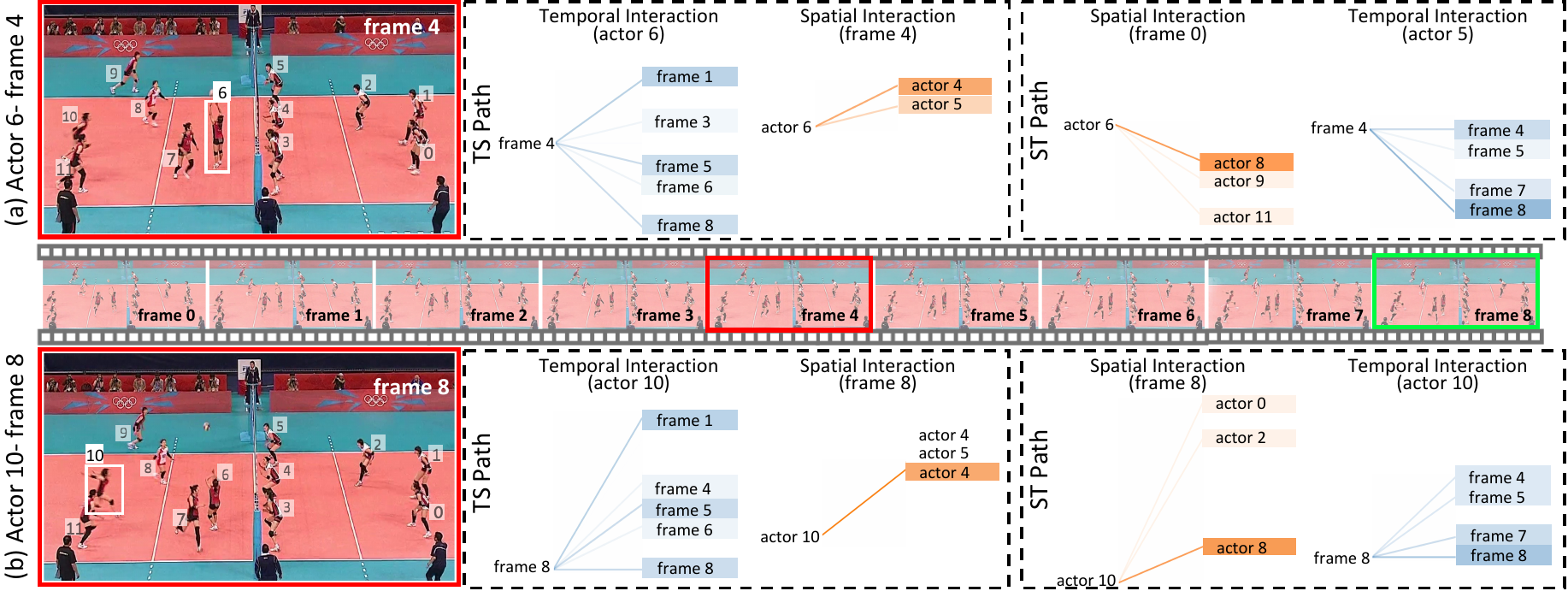}
  \captionof{figure}{
      Actor interaction visualization for \textit{l-set} activity with connected lines.
      Brighter color indicates stronger relation.
      (a) For actor 6 in frame 4, we visualize the temporal interaction with same actors in different frames for ST and TS paths; 
      similarly,
      we visualize the spatial interaction with different actors in frame 4.
      (b) We visualize the actor interaction for actor 10 in frame 8 in the same way.
  }
  \label{fig:supp_vis_attention1}
\end{center}%
}]

\appendix
    
    \section{Implementation Details}
    For Vollyball and Weak-Volleyball-M, 
    we randomly select $K\!=\!3$ frames with 720$\times$1280 resolution for training and 9 frames for testing, 
    corresponding to 4 frames before the middle frame and 4 frames after. 
    For Collective Activity dataset, 
    we utilize $K\!=\!10$ frames (480$\times$720) of each video clip for training and testing.
    For NBA dataset, 
    we select $K\!=\!3$ frames (720$\times$1280) around middle frame of each video for training and take 20 frames for testing. 
    For Volleyball and Collective Activity dataset,
    we use annotated bounding boxes provided by the datasets for training and testing to make fair comparison,
    \ie,
    $N=12$ and $N=13$ respectively.

    \textbf{Optimization.} We adopt Adam \cite{kingma2014adam} to learn the network parameters with initial learning rate set to 0.0001.
    We run 140 epochs in total to obtain the reported results and decay the learning rate by 10 after 60 and 100 epochs.
    We implement our method based on the released code\footnote{https://github.com/wjchaoGit/Group-Activity-Recognition} of \cite{wu2019learning} and transformer code\footnote{ v1.6.0/torch/nn/modules/transformer.py} from Pytorch.
    
    \textbf{Weakly Supervised GAR.} We detect actors in the centered frame of each clip with MMDetection Toolbox \cite{mmdetection}.
    Following \cite{yan2020social},
    we further obtain the tracklets by correlation tracker \cite{danelljan2014accurate} implemented by Dlib \cite{king2009dlib}.
    Specificly,
    we use a Faster-RCNN \cite{ren2015faster} with ResNet-50 \cite{he2016deep} backbone provided by MMDetection, 
    which is pretrained on COCO dataset \cite{lin2014microsoft} and further finetuned with person subset of COCO. 
    We use the default configuration provided and sort the detected boxes by the confidence scores.
    Finally,
    we select the top $N$ (16 for NBA Dataset and 20 for Weak-Volleyball-M) bounding boxes for actor interaction learning.
    
    \textbf{Limited GAR.}
    We randomly select videos in Volleyball dataset for limited setting.
    The video ids used for 5\%, 10\%, 25\% and 50\% data in our experiments are (1, 38), (1, 23, 38, 54), (1, 6, 10, 15, 18, 23, 32, 38, 42, 48) and (1, 3, 6, 7, 10, 13, 15, 16, 18, 22, 23, 31, 32, 36, 38, 39, 40, 41, 42, 48), respectively.
    We implement results of other methods with the released code and configurations from \cite{yuan2021spatio}.
    
    \section{Dataset}
    \label{sec:suppl_dataset}
    We provide more details of the datasets used for the convenience of result production.
    
    \noindent
    \textbf{Volleyball Dataset.} 
    Each clip is annotated with one of 8 group activity classes:
    \textit{right set, right spike, right pass, right win-point,
    left set, left spike, left pass and left win-point}. 
    Middle frame of each clip is annotated with 9 individual action labels
    (\textit{waiting, setting, digging, falling, spiking, blocking, jumping, moving and standing})
    and their bounding boxes.
    
    \noindent
    \textbf{Collective Activity Dataset.}
    The individual is annotated with one of the following 6 action categories 
    (\textit{NA, crossing, waiting, queuing, walking and talking}) 
    and their bounding boxes. 
    We follow \cite{yan2020higcin,yan2018participation,yuan2021spatio} to merge the \textit{crossing} and \textit{walking} into \textit{moving}.
    Train-test split follows \cite{yuan2021spatio}.
    
    \noindent
    \textbf{Weak-Volleyball-M.} 
    No standalone distribution is released.
    This dataset \cite{yan2020social} is adapted from Volleyball dataset \cite{ibrahim2016hierarchical},
    merging \textit{pass} and \textit{set} into one category and discarding all individual annotations.

    \noindent
    \textbf{NBA Dataset.} 
    This dataset is available upon request\footnote{https://ruiyan1995.github.io/SAM.html},
    limited by the copyright.
    Each video clip is annotated with one of the 9 group activities:
    \textit{2p-succ., 2p-fail.-off.,	
    2p-fail.-def.,
    2p-layup-succ.,	
    2p-layup-fail.-off.,	
    2p-layup-fail.-def.,
    3p-succ.,	
    3p-fail.-off.,	
    3p-fail.-def.}.
    No individual annotations, 
    such as individual action labels and bounding boxes, 
    are provided.
    
    \section{Visualization}
    We provide visualization of actor interactions for \textit{l-set},
    as shown in \cref{fig:supp_vis_attention1}.
    The attention weight between actors is represented by connected lines, 
    and the brightness of the lines represents the scale of the attention weight.
    Orange and Blue lines correspond to the Spatial and Temporal interaction,
    respectively.
    
    As shown by spatial interaction in \cref{fig:supp_vis_attention1} (a), 
    the player setting the ball (actor 6) is more related with defending players in TS path,
    who are ``jumping'' and ``blocking'' (actor 4 and actor 5).
    Differently,
    in ST path,
    actor 6 has wider connections with accompanying players,
    who are ``moving'' (actor 8 and actor 9) and ``jumping'' (actor 10) cooperatively to tackle the ball falling on different position and prepare for next ball contact.
    Similarly,
    as shown by spatial interaction in \cref{fig:supp_vis_attention1} (b),
    the actor 10 is related to defending player (actor 4) in TS path,
    and related to both accompanying player (actor 8) and defending players (actor 0 and actor 2),
    showing complementary patterns.
    As for temporal interaction in both (a) and (b),
    the anchor actor is more evenly related to other frames (frame 1, frame 5 and frame 8) in TS path,
    and more related to late frames (frame 7 and frame 8) in ST path,
    which shows different evolution patterns.
    

    \end{document}